\documentclass[10pt,twocolumn,letterpaper]{article}
\pdfoutput=1
\usepackage{iccv}              % To produce the CAMERA-READY version
\usepackage{times}
\usepackage{epsfig}
\usepackage{graphicx}
\usepackage{amsmath}
\usepackage{amssymb}
\usepackage{textcomp}
\usepackage{color}
\usepackage{subcaption}
\usepackage{booktabs}
\usepackage{tabularx}
\usepackage{multirow}
\usepackage{float}  %设置图片浮动位置的宏包
\usepackage{caption}
\usepackage{float} 
\usepackage{overpic} % 图片内引用

\ifodd 1
\newcommand{\com}[1]{\textbf{\color{red}(COMMENT: #1)}} %comment of the 
\else

\newcommand{\com}[1]{}
\fi

\newcommand{\model}{PRE-Mamba}
\newcommand{\datasetall}{27K}
\newcommand{\datasetsyn}{7K}
\newcommand{\datasetart}{7K}
\newcommand{\datasetreal}{9K}
\newcommand{\datasetlab}{18K}
\newcommand{\datasetsnow}{4K}
\definecolor{iccvblue}{rgb}{0.21,0.49,0.74}
\usepackage[pagebackref,breaklinks,colorlinks,allcolors=iccvblue]{hyperref}

\def\paperID{1614} % *** Enter the Paper ID here
\def\confName{ICCV}
\def\confYear{2025}

\title{
    \includegraphics[height=0.8cm]{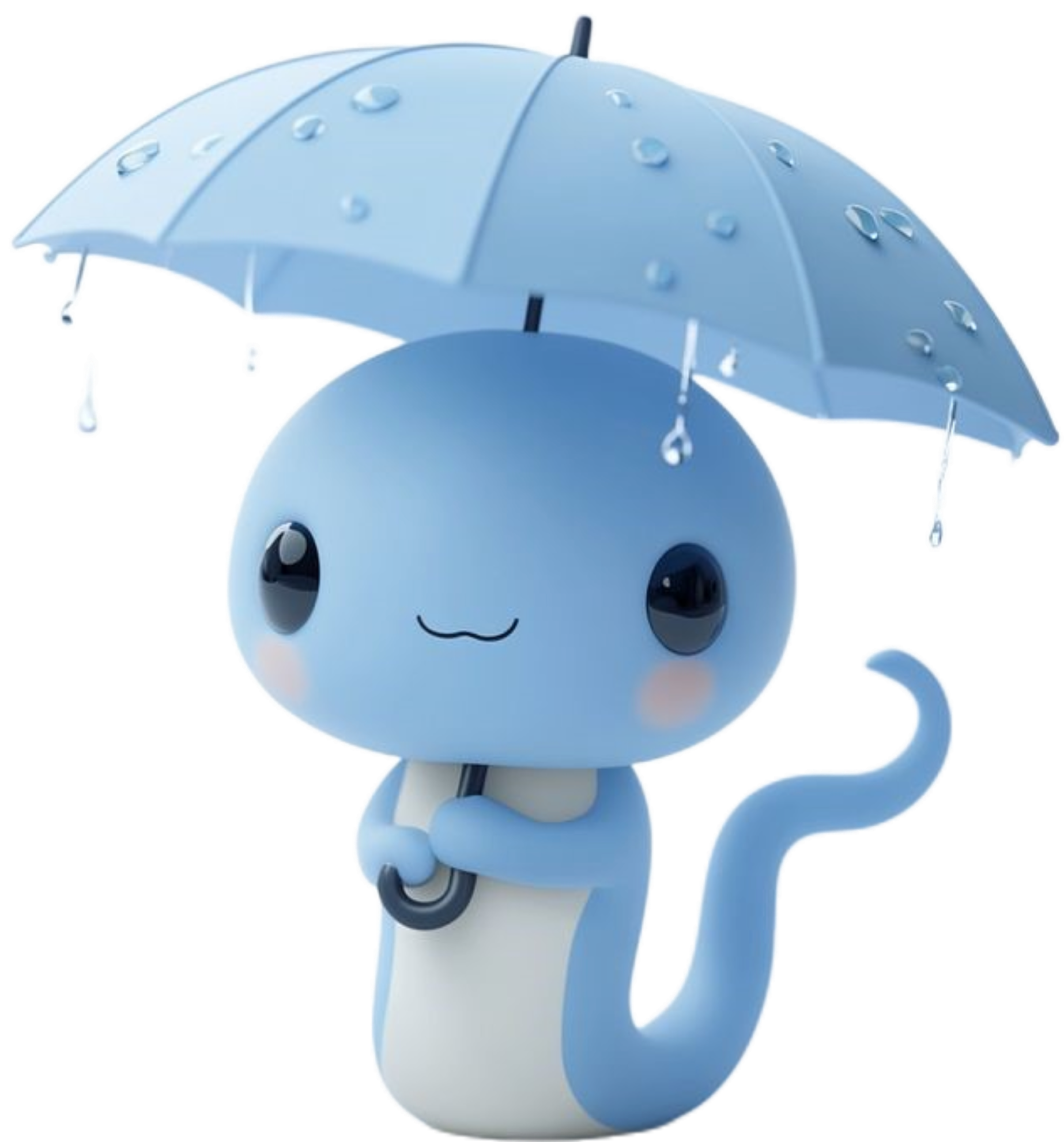}
    \ \model: A 4D State Space Model for Ultra-High-Frequent \\Event Camera Deraining
    }
\usepackage{stfloats}
\author{
Ciyu Ruan\textsuperscript{1,*}, 
Ruishan Guo\textsuperscript{1,*}, 
Zihang Gong\textsuperscript{2}, 
Jingao Xu\textsuperscript{3},
Wenhan Yang\textsuperscript{4}, 
Xinlei Chen\textsuperscript{1, \dag}  \\
\textsuperscript{1}Shenzhen International Graduate School, Tsinghua University,
\textsuperscript{2}Harbin Institute of Technology,\\
\textsuperscript{3}Carnegie Mellon University,
\textsuperscript{4}Pengcheng Laboratory,\\
{\tt\small \{rcy23, grs24\}@mails.tsinghua.edu.cn, 
gongzihang0201@gmail.com}, \\
{\tt\small jingaox@andrew.cmu.edu}, \ {\tt\small yangwh@pcl.ac.cn}, \ {\tt\small chen.xinlei@sz.tsinghua.edu.cn}
}

\begin{document}
\twocolumn[{%
\renewcommand\twocolumn[1][]{#1}%
\maketitle
\centering
\vspace{-0.8cm}
\begin{overpic}[width=0.96\linewidth]{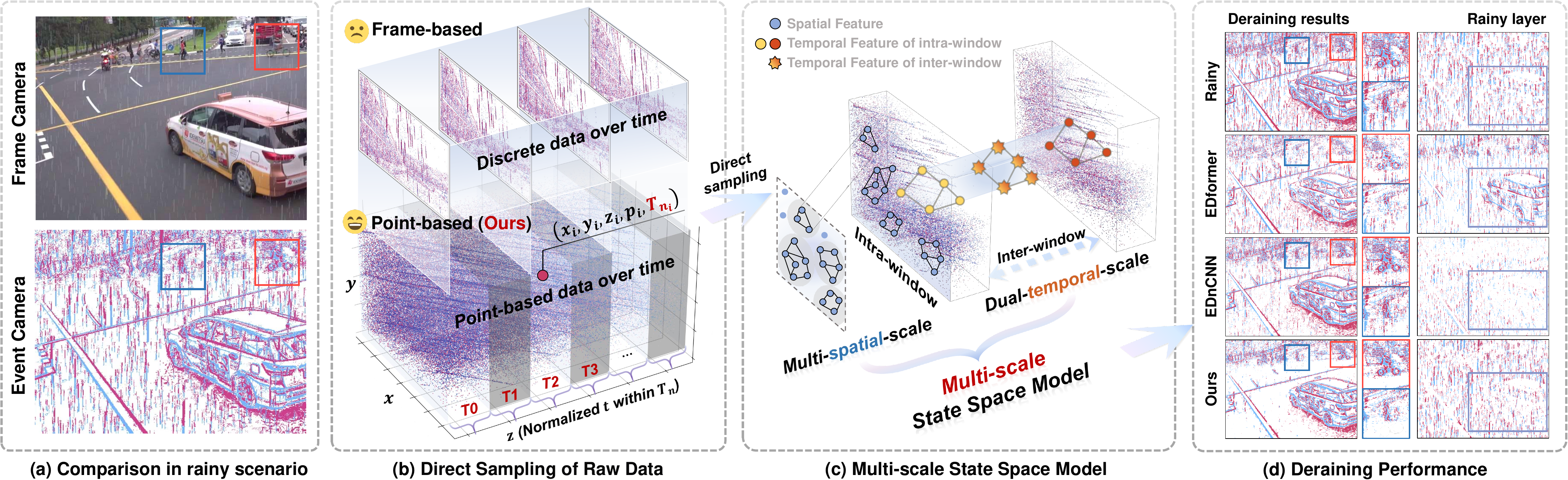}
    \put(40.7,29){\color{red}\scriptsize\textbf{Sec.~\ref{sec:4d}}}
    \put(66.5,29){\color{red}\scriptsize\textbf{Sec.~\ref{sec:stdf}}}
    \put(71.5,29){\color{red}\scriptsize\textbf{,~\ref{sec:ms3m}}}
    \put(88,29){\color{red}\scriptsize\textbf{Sec.~\ref{sec:experiment}}}
\end{overpic}
\vspace{-0.9em}
% \includegraphics[width=1.0\linewidth,height=0.4\linewidth]{ICCV2025-Author-Kit-Feb/img/teasor_v3.png}
% \includegraphics[width=.32\linewidth]{example-image-b.png}
% \includegraphics[width=.32\linewidth]{example-image-c.png}
% \vspace{-2em}
\captionof{figure}{(a) Rain severely degrades event camera data compared to conventional cameras. (b) Unlike frame-based aggregation methods, our approach directly samples raw data, preserving temporal resolution. (c) The multi-scale state space model integrates dual-temporal scales for intra- and inter-window information and multi-spatial scales for rain dynamics. (d) Deraining results of our method.}
\vspace{0.6em}
\label{fig:teaser}
}]
\renewcommand{\thefootnote}{\fnsymbol{footnote}}
\footnotetext[1]{These authors contributed equally.}
\footnotetext[2]{Corresponding author.}
\renewcommand{\thefootnote}{\arabic{footnote}}  % 恢复正常脚注编号样式

\begin{abstract}
Event cameras excel in high temporal resolution and dynamic range but suffer from dense noise in rainy conditions.
 Existing event deraining methods face trade-offs between temporal precision, deraining effectiveness, and computational efficiency. 
 In this paper, we propose \model, a novel point-based event camera deraining framework that fully exploits the spatiotemporal characteristics of raw event and rain. 
 Our framework introduces a 4D event cloud representation that integrates dual temporal scales to preserve high temporal precision, 
a Spatio-Temporal Decoupling and Fusion module (STDF) that enhances deraining capability by enabling shallow decoupling and interaction of temporal and spatial information, 
and a Multi-Scale State Space Model (MS3M) that captures deeper rain dynamics across dual-temporal and multi-spatial scales with linear computational complexity.
Enhanced by frequency-domain regularization, \model \ achieves superior performance (0.95 SR, 0.91 NR, and 0.4s/M events) with only 0.26M parameters on EventRain-\datasetall, a comprehensive dataset with labeled synthetic and real-world sequences. 
Moreover, our method generalizes well across varying rain intensities, viewpoints, and even snowy conditions. Code and dataset: https://github.com/softword-tt/PRE-Mamba.
\end{abstract}    
\vspace{-0.5cm}
\section{Introduction}
\label{sec:intro}
Event cameras have revolutionized mobile applications with their high temporal resolution (microsecond level), high dynamic range (\textgreater 120 dB), and low power consumption (\textless 10 mW)~\cite{gallego2020event,wang2025towards}, driving breakthroughs in high-speed tracking~\cite{glover2017robust, luo2024eventtracker}, SLAM~\cite{jiao2021comparing,vidal2018ultimate,gehrig2024low}, localization~\cite{wang2025ultra} and obstacle avoidance~\cite{falanga2020dynamic,sanket2020evdodgenet}. Unlike frame cameras, they asynchronously capture pixel-level intensity changes without global exposure, eliminating motion blur even at extreme speeds. 
However, despite excelling in dynamic scenes, event cameras suffer significant performance degradation under adverse weather, particularly in rain~\cite{ruan2024distill,cheng2022novel}. 
The rapid motion of rain streaks triggers excessive intensity changes, generating dense noise that overwhelms valid scene information and exacerbates sampling rates. This severely impacts accuracy and latency in downstream tasks on drones~\cite{chen2015drunkwalk,zhang2019cellular,wang2022h} and autonomous 
vehicles~\cite{jian2023path,messaoudi2024ugv,jianzhuozhu2025a}, limiting real-world reliability.

Despite significant advances in deep learning for frame camera deraining~\cite{yu2021single,chen2023learning, karavarsamis2022survey,huang2024progressive}, event camera deraining remains largely unexplored. 
Existing solutions predominantly follow a frame-based paradigm, converting event streams into grayscale frames~\cite{cheng2022novel} and applying frame-based deraining techniques~\cite{ruan2024distill}, which sacrifice the fine-grained temporal resolution and sparsity of event data. Point-based methods align naturally with event streams' asynchronous, sparse nature, but remain unapplied to deraining due to notable limitations.
% Point Networks~\cite{wang2019space,sekikawa2019eventnet}, GNNs~\cite{bi2020graph,schaefer2022aegnn}, and SNNs~\cite{kugele2021hybrid, schnider2023neuromorphic, 9711070} still underperform compared to frame-based approaches, while Point Transformers struggles with computational bottleneck~\cite{diazmamba,ren2024rethinking} under high event rates (\textgreater 10M events/s). As a result, achieving a balance between temporal precision, deraining effectiveness, and computational efficiency remains an open challenge.
Point Networks~\cite{wang2019space,sekikawa2019eventnet}, GNNs~\cite{bi2020graph,schaefer2022aegnn}, and SNNs~\cite{kugele2021hybrid,schnider2023neuromorphic,9711070} still underperform compared to frame-based methods, while Point Transformers face computational bottlenecks under high event rates (\textgreater 10M events/s) due to quadratic attention complexity~\cite{diazmamba,ren2024rethinking}. As a result, balancing temporal precision, deraining effectiveness, and computational efficiency remains an open challenge.

Recently, the State Space Model (SSM)~\cite{gu2023mamba}, particularly Mamba, provides a promising solution with linear complexity and long-range context modeling. However, introducing Mamba to event camera deraining is non-trivial. The original design of the Mamba is aimed at solving causal sequential language tasks, which differ from the asynchronous, sparse, and high-temporal-resolution nature of event data.  Rain streaks further complicate this by introducing diverse spatiotemporal patterns, requiring the model to capture complex dependencies across both time and space.

% To address this,  we propose \textbf{\model}, a novel \textbf{P}oint-based de\textbf{R}aining framework for  \textbf{E}vent camera. Our framework first introduces a 4D event cloud representation, integrating inter- and intra-temporal windows to preserve high temporal precision. To handle the spatial irregularity and motion characteristics of rain, we propose a Spatio-Temporal Decoupling and Fusion module (STDF), enabling shallow decoupling and interaction of temporal and spatial information. For deeper modeling, we propose a Multi-Scale State Space Model (MS3M) that efficiently captures complex rain dynamics across dual temporal and multi-spatial scales with linear computational complexity. To improve feature discrimination, we introduce a frequency regularization term in the loss function, guiding the model to learn rain distribution patterns in the frequency domain.

To address this,  we propose \textbf{\model}, the first \textbf{P}oint-based de\textbf{R}aining framework for  \textbf{E}vent camera. 
As a point-based method, it processes raw events directly instead of quantized frames, fully preserving the sensor's native microsecond (µs) temporal resolution and facilitating downstream high-speed vision tasks.
Our framework first introduces a 4D event cloud representation (\S\ref{sec:4d}), integrating inter- and intra-temporal windows to preserve high temporal precision. To handle the spatial irregularity and motion characteristics of rain, we propose a Spatio-Temporal Decoupling and Fusion module (STDF, \S\ref{sec:stdf}), enabling shallow decoupling and interaction of temporal and spatial information. For deeper modeling, we propose a Multi-Scale State Space Model (MS3M, \S\ref{sec:ms3m}) that efficiently captures complex rain dynamics across dual temporal and multi-spatial scales with linear computational complexity. To improve feature discrimination, we introduce a frequency regularization term (\S\ref{sec:loss}) in the loss function, guiding the model to learn rain distribution patterns in the frequency domain.
To evaluate our approach, we construct EventRain-\datasetall \  (\S\ref{sec:dataset}), the first comprehensive point-based event deraining dataset, comprising \datasetlab \  labeled synthetic and \datasetreal \ unlabeled real-world rain sequences.  Our model achieves state-of-the-art performance (0.95 SR, 0.91 NR, and 0.4s/M events) with only 0.26M parameters, validated through extensive experiments (\S\ref{sec:experiment}). Moreover, 
% it generalizes well across varying rain intensities, viewpoints, and even snowy conditions.
it generalizes well across diverse rain intensities and snowy conditions.
Our contributions are summarized as follows:
\begin{itemize}
    \item  To the best of our knowledge,  we propose the first point-based event deraining framework.
    % \item We extend Mamba for event deraining by introducing a 4D event cloud representation, a Spatio-Temporal Decoupling and Fusion module, and a Multi-Scale State Space Model, enhanced with a frequency regularization loss.
    \item To effectively extend Mamba for event deraining, we address the asynchronous, sparse nature of event data and the diverse spatiotemporal patterns of rain streaks by introducing a 4D event cloud, an STDF module, and an MS3M enhanced with frequency regularization loss.
    \item We introduce EventRain-\datasetall, the first point-based event deraining with labeled synthetic and real-world sequences, providing a benchmark for future research.
    \item We achieve superior performance on synthetic and real datasets with high efficiency and a lightweight design.
\end{itemize}

\section{Related Work}
\label{sec:related}

\noindent \textbf{Existing deraining approaches for frame camera.}
Deraining methods for frame cameras are categorized into traditional and deep learning-based approaches. Traditional methods, such as dictionary learning and low-rank matrix decomposition~\cite{https://doi.org/10.1049/iet-ipr.2018.5122, Wang2020AMD, 7101234, Chang_2017_ICCV}, rely on handcrafted assumptions about the directional nature or sparsity of rain streaks, often failing to generalize to complex real-world scenarios. In contrast, deep learning-based methods, leveraging CNNs and Transformer~\cite{liang2021swinir,Wang_2022_CVPR} architectures, have shown significant advances. CNNs effectively model local spatial patterns, capturing rain stripe textures and background structures~\cite{yu2021single, Zamir_2021_CVPR}, 
% while Transformer-based models like SwinIR~\cite{liang2021swinir} and Uformer~\cite{Wang_2022_CVPR}  enhance performance by capturing long-range dependencies and global context, offering superior deraining capabilities.
while Transformer-based models~\cite{huang2024progressive, chen2023learning, karavarsamis2022survey} enhance performance by capturing long-range dependencies and global context, offering superior deraining capabilities.

% \subsection{Existing derain approaches for event camera}
% \noindent \textbf{Existing derain approaches for event camera.} Though rain removal is crucial for outdoor applications of event cameras, research on event-based deraining remains relatively limited. 
% Existing event-based deraining methods predominantly utilize the high temporal resolution and dynamic range of event cameras to assist frame cameras in mitigating rain effects. EGVD\cite{Zhang2023EGVDEV} , RainVID\&SS\cite{sun2023event} and NESD\cite{ge2024neuromorphic} integrate event data to provide complementary modality information and prior knowledge of the rain location and density.

\noindent \textbf{Existing deraining approaches involving event camera.}
Research on event-based deraining remains limited despite its importance for outdoor applications. 
Existing methods, such as EGVD~\cite{Zhang2023EGVDEV}, RainVID\&SS~\cite{sun2023event}, and NESD ~\cite{ge2024neuromorphic}, leverage the high temporal resolution and dynamic range of event cameras to assist frame camera by providing prior knowledge of rain location and density.
Other methods designed for event-based deraining typically convert asynchronous event streams into synchronous frame-like representations, aligning with image deraining but sacrificing temporal resolution and sparsity. 
For instance, DistillNet~\cite{ruan2024distill} partially preserves temporal resolution through voxel representations but still loses their asynchronous nature. 
Moreover, the unique motion and spatiotemporal distribution of rain create a domain barrier, limiting the applicability of traditional event denoising methods like filtering~\cite{duan2021guided, app10062024, 10181865} and clustering~\cite{10138453}.
% and others~\cite{9720086, baldwin2020event, 10078400,9578367}.
In addition, existing deraining datasets primarily focus on frame camera~\cite{wang2019spatial, hu2019depth, yang2020single, wang2021rain} or represent event data in image-like formats~\cite{10377341}, further hinder the event camera deraining research.

% This often results in blurred rain streaks and background occlusion. 
% For instance, DistillNet~\cite{ruan2024distill} partially preserves temporal resolution through voxel representations but still discretizing event streams and losing their asynchronous nature.
% Apart from these, methods specifically designed for event-based deraining typically convert event data into image formats \cite{Chen2022ANR, 10760025}, aligning with traditional techniques but sacrificing temporal resolution and sparse data advantages\cite{ren2024rethinking}. \cite{ruan2024distill} uses a hybrid SNN-CNN network to separate rain and background events. While their voxel representation partially preserves temporal resolution, it remains frame-like as it discretizes event streams into fixed bins, losing the asynchronous nature of raw event data.

% In addition, due to the unique motion and spatiotemporal distribution of rain, there is a domain barrier between noise and rain, which leads to the limited applicability and insufficient constraint of filtering\cite{duan2021guided, app10062024, 10181865}, clustering\cite{10138453}, and other traditional event denoising methods\cite{9720086, baldwin2020event, 10078400,9578367}.

% \subsection{Event Data Processing Architectures}

\begin{figure*}[htb]
  \centering
% \fbox{\rule{0pt}{3in} \rule{0.9\linewidth}{0pt}}
   \includegraphics[width=1.0\linewidth]{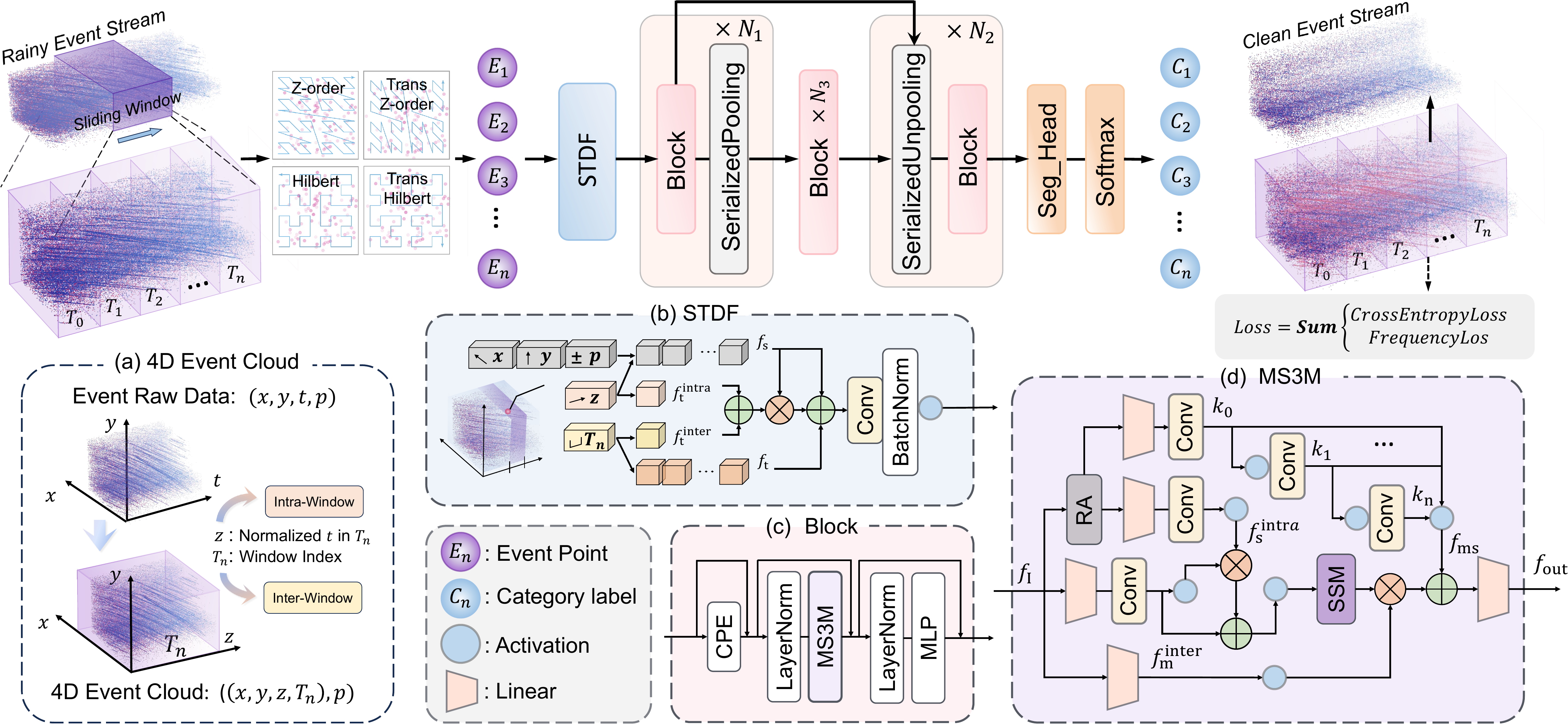}
    \vspace{-0.6cm}
  \caption{\textbf{ Overview of Our Proposed \model.} The model takes $n$ temporal windows as input to generate a 4D event cloud, serialized via four scanning modes. The STDF module decouples and enhances spatiotemporal features across dual temporal scales, which flow through a U-Net style encoder-decoder with multiple blocks.
  Each block employs MS3M as its core, integrating intra- and inter-window branches to capture appearance and motion features, alongside a multi-spatial-scale pathway for handling diverse rain streak scales while preserving local dynamics.  Finally, the decoder outputs are fed into a linear layer followed by softmax for per-event predictions.}
  \vspace{-0.6cm}
\label{fig:pipeline}
\end{figure*}

\noindent \textbf{Event Data Processing Architectures.} 
Event data processing approaches are broadly categorized into frame-based and point-based frameworks.
Frame-based methods~\cite{8578884, gehrig2019end,10181865, 9191148} not only face the aforementioned limitations but also incur redundant computational overhead by processing event-free regions~\cite{Zubic_2024_CVPR}, making them inefficient for resource-constrained environments~\cite{gehrig2019end, 9412991}.
In contrast, point-based methods process events in their native form, preserving temporal resolution and sparsity.  These include Point Networks~\cite{NEURIPS2023_8296d580,qi2017pointnet}, Graph Neural Networks (GNNs)~\cite{Gehrig24nature}, and Spiking Neural Networks (SNNs)~\cite{9711070, shen2023esl}, excel in capturing rapid motion and fine-grained temporal dynamics.  
However, GNNs and SNNs often require specialized hardware or face performance issues, while Point Cloud networks, based on assumptions of permutation and transformation invariance, are poorly suited for temporal event data and require time-aware feature extraction.

To address these limitations, this paper proposes a novel approach based on selective State Space Models (SSMs)~\cite{zhen2024freqmamba, guo2024mambair, zhu2024vision}.
Renowned for its linear computational complexity and parameter efficiency~\cite{gu2023mamba, sun2024hybrid}, Mamba is well-suited for processing long event sequences exceeding 1M length~\cite{diazmamba}.
It provides a promising direction to bridge performance gap between point-based and frame-based methods~\cite{diazmamba, li2025multi} while preserving the inherent advantages of event data, facilitating the development of lightweight, real-time deraining systems.

\vspace{-0.3cm}
\section{Method}
\label{sec:method}
To perform event deraining effectively and efficiently while preserving temporal precision, we propose a novel framework, \model. We formulate event deraining as an event-by-event classification task, akin to event denoising, reducing computational complexity and avoiding artifacts compared to reconstruction-based methods. Figure \ref{fig:pipeline} illustrates the architecture of our system. At the input level, a 4D Event Cloud Representation (\S\ref{sec:4d}) integrates inter- and intra-temporal windows to preserve high temporal precision. For feature extraction, a Spatio-Temporal Decoupling and Fusion module (\S\ref{sec:stdf}) enables effective decoupling and interaction of temporal and spatial information to capture complex rain dynamics. For deep modeling, a Multi-Scale State Space Model (\S\ref{sec:ms3m}) captures rain dynamics across dual temporal and multi-spatial scales with linear computational complexity. Finally, a Frequency-Regularized Optimization (\S\ref{sec:loss}) enhances feature discrimination by learning rain distribution patterns in the frequency domain, ensuring robust deraining accuracy. Together, these components form a unified framework that achieves both computational efficiency and deraining performance.

\subsection{4D Event Cloud Representation}
\label{sec:4d}
To preserve fine-grained spatiotemporal information while enabling efficient processing, we transform raw event streams into a structured 4D event cloud representation. The raw event data is represented as a sequence $\mathcal{E} = \{e_i\}_{i=1}^N$, where each event $e_i = (x_i, y_i, t_i, p_i)$  consists of spatial coordinates $(x_i, y_i)$, timestamp $t_i$, and polarity $p_i$.

\noindent \textbf{3D Pseudo-Point Cloud Construction:}  
The event sequence is divided into fixed-duration temporal windows $\{\mathcal{W}_k\}_{k=1}^L$, each spanning a time interval $T$. Within each window $\mathcal{W}_k$, events are normalized into a 3D pseudo-point cloud representation:
\vspace{-0.3cm}
\begin{equation}
p_i = (x_i, y_i, z_i) = \left(x_i, y_i, \frac{t_i-t_0}{t_e-t_0}\right),
\end{equation}
where $t_0$ and $t_e$ denote the window's start and end timestamps, respectively. The normalized coordinate $z_i$ encodes the relative temporal position of the event within the window, mapping event timestamps to a unified temporal scale.

\noindent \textbf{4D Cross-Window Temporal Modeling:}  
To model long-term dependencies across windows, we introduce a fourth dimension, $T_n$, which represents the global temporal index of the window.  This extends the 3D representation to a 4D event cloud ${E}_n = \{(x_i, y_i, z_i, T_n)\}_{i=1}^{M_n}$,
where $M_n$ is the number of events in window $n$. The complete 4D event cloud is then ${E} = \bigcup_{n=1}^L {E}_n$.

% The conventional frame-based approach aggregates events over fixed intervals, inevitably sacrificing temporal resolution. In contrast, our hierarchical representation preserves intrinsic temporal precision through a dual-scale framework: local intra-window dynamics are captured via $z_i$
%  , while global inter-window relationships are modeled via $t_k$. This enables microsecond-level resolution and long-range dependencies across scales, achieving comprehensive spatiotemporal modeling while maintaining compatibility with point cloud pipelines.
% Conventional frame-based methods aggregate events over fixed intervals, sacrificing temporal resolution and may generate blurred images. 
This hierarchical dual-scale framework captures local intra-window dynamics through normalized timestamp $z$ and models global inter-window evolution via window index $T_\text{n}$. 
% Some point-based event denoising methods~\cite{ren2024rethinking,jiang2024edformer} enforce a fixed event count $N$, fragmenting or regrouping events and compromising temporal integrity.  
% In contrast, our approach leverages a state space model (SSM) to dynamically adapt to event sparsity and irregularity by processing variable-length sequences within fixed time windows. 
% In contrast, our approach dynamically adapts to event sparsity and irregularity by processing variable-length sequences within fixed time windows. 
%  Additionally, we serialize 4D event clouds into SSM-compatible sequences using z-order and Hilbert curves, aligning with point cloud processing paradigms~\cite{zhao2021point,wu2022point}.
 Unlike fixed event-count methods that fragment or regroup events~\cite{ren2024rethinking,jiang2024edformer}, compromising temporal integrity, our approach dynamically adapts to event sparsity and irregularity by processing variable-length sequences within fixed time windows. We also serialize 4D event clouds into SSM-compatible sequences using z-order and Hilbert curves, aligning with point cloud processing paradigms~\cite{zhao2021point,wu2022point}.

\subsection{Spatio-Temporal Decoupling and Fusion}
\label{sec:stdf}
To address the complex spatiotemporal noise in rain events while maximizing the temporal discrimination power of event camera, we propose a Spatio-Temporal Decoupling and Fusion Module (STDF). This module explicitly decouples spatial and temporal features while modeling their interactions across dual scales in the 4D event cloud.
% , enabling more effective noise filtering and signal enhancement in dynamic rainfall scenarios.

We extract spatial features $f_\text{s}$ via depthwise 1D convolutions on $(x, y, z, p)$ coordinates and temporal features $f_\text{t}$ through embeddings of window indices $T_\text{n}$. To exploit rain events' temporal characteristics and event cameras' high resolution, we prioritize temporal information over spatial information through cross-domain feature modulation. Specifically, spatial features $f_\text{s}$ are refined through Hadamard products with complementary temporal representations:  
intra-window dynamics $f_{t}^{\text{intra}}$ from 1D convolutions along $z$, encoding microsecond correlations within windows, and inter-window trends $f_{t}^{\text{inter}}$ from 1D convolutions on $T_\text{n}$, capturing long-term temporal trends across consecutive windows. 
This modulation is formulated as:
\vspace{-0.1cm}
\begin{equation}
    f_\text{s}^{*} = f_\text{s} \otimes \left( f_\text{t}^{\text{intra}} + f_{t}^{\text{inter}} \right).
\end{equation}

Finally, the temporally-modulated features are combined with the original spatial and temporal representations via residual addition ($\mathcal{R}$). The integrated features are  sequentially processed by 1D convolution ($\Phi_{1d}$), batch normalization ($\text{BN}$), and activation ($\sigma$) to ensure training stability:
\vspace{-0.2cm}
\begin{equation}
\begin{aligned}
\mathcal{R}(f_\text{s}, f_\text{t}) &= f_\text{s} + f_\text{t} + f_\text{s}^{*}, \\
f_{\text{STDF}} &= \sigma \left( \text{BN} \left( \Phi_{1d} \left( \mathcal{R}(f_\text{s}, f_\text{t}) \right) \right) \right).
\end{aligned}
\end{equation}

\subsection{Multi-Scale State Space Model}
\label{sec:ms3m}
Building upon the hierarchical input representation and spatiotemporal feature embedding, we propose the Multi-Scale State Space Model (MS3M) to further model deeper rain dynamics from multi-temporal and spatial scales.
This framework incorporates parallel intra- and inter-window branches to simultaneously capture appearance features and motion patterns, coupled with a multi-scale spatial pathway that preserves structural details and local rain dynamics. 

\noindent \textbf{Dual-Temporal-Scale Global Modeling:} 
While Mamba excels in sequence modeling, it struggles to leverage temporal features~\cite{qin2024mamba} for event deraining. Inspired by the Motion-aware State Space Model (MSSM), which excels in LiDAR-based moving object segmentation~\cite{zeng2024mambamos}, we integrate MSSM into our dual-temporal-scale architecture. This enables explicit modeling of rain events' appearance and motion, improving rain–background separation.

The architecture mainly consists of two complementary branches: an intra-window branch processes spatial appearance features $f_{\text{s}}^{\text{intra}}$ through Reversed Aggregation (RA) and 1D convolution, capturing rain patterns and structural details for rain-background separation; an inter-window branch extracts motion-aware features $f_{\text{m}}^{\text{inter}}$ by analyzing temporal dependencies, enabling separation of dynamic rain streaks from moving objects or scene changes. 
An adaptive gating mechanism~\cite{hua2022transformer} generates weight $f_\text{G}$ to emphasize rain-relevant features.
Outputs from both branches are fused via cross-product attention with inter-branch interaction, then passed to SSM for global feature learning via scanning~\cite{zhang2024cdmamba}, followed by gating signal modulation.
% Features from both branches are fused via cross-product attention, it is sequentially fed into Conv1d layer and SSM to learn global features through scanning, and modulated through element-wise multiplication with gating signals.
% The intra-window branch focuses on appearance features $f_{intra}$ within individual windows through Reversed Aggregation (RA)\tocite and 1D convolution, while the inter-window branch captures the dynamic motion patterns and extracts motion-aware features $f_{\text{inter}}$ by analyzing rain behavior across multiple temporal windows. To dynamically prioritize rain-related features over background features, we introduce a gating mechanism that assigns adaptive weights $f_G$ to different feature components. The features from both branches are fused through cross-product attention, then fed into an SSM (State Space Model), and further modulated by element-wise multiplication with the gate branch:
% \begin{equation}
%     \begin{aligned}
%        f_{{s}}^{\text{intra}} &= \sigma(\Phi_{1d}(\text{RA}(f_I))) \\
%         f_{\text{fuse}} &= \sigma(f_{{m}}^{\text{inter}}) \otimes f_{{s}}^{\text{intra}} + f_{{m}}^{\text{inter}} \\
%         f_{\text{dual}}  &= \text{SSM}(\sigma(f_{\text{fuse}})) \otimes f_G .
%     \end{aligned}
% \end{equation}  
\vspace{-0.2cm}
\begin{equation}
    \begin{aligned}
       f_{\text{s}}^{\text{intra}} &= \sigma(\Phi_{1d}(\text{RA}(f_\text{I}))) \\
        f_{\text{fuse}} &= \sigma(f_{\text{m}}^{\text{inter}}) \otimes f_{\text{s}}^{\text{intra}} + f_{\text{m}}^{\text{inter}} \\
        f_{\text{dual}}  &= \text{SSM}(\sigma(f_{\text{fuse}})) \otimes f_\text{G} .
    \end{aligned}
\end{equation}  

% This design facilitates a comprehensive understanding of rain dynamics by simultaneously analyzing dual temporal resolutions and integrating both appearance and motion characteristics.

% The RSSM processes event data through a dual-branch architecture, where each branch is responsible for different aspects of the input. The single-window branch extracts spatial appearance features from individual event windows by applying 3D convolutions over the $(x,y,z)$ coordinates, which capture local geometric structures and polarity patterns. 
% Meanwhile, the multi-window branch captures temporal dynamics across multiple event windows, focusing on the temporal features of moving objects, particularly rain, enabling the model to differentiate rain artifacts from valid event patterns and helping capture long-range dependencies and motion continuity.
%  The spatial features from the single-window branch are modulated by the temporal features from the multi-window branch, allowing the model to focus on the most relevant temporal context for each spatial element. This fusion enables the model to effectively capture both local geometric details and long-range temporal dependencies, resulting in a comprehensive representation of the event data.

\noindent \textbf{Multi-Spatial-Scale Local Pathway:}
% To complement the global modeling branches,   While the intra-window and inter-window branches capture global appearance and motion patterns, they may overlook subtle spatial variations critical for high-quality restoration. 
To address rain scale and appearance diversity, 
% To further enhance the representation of rain streaks and address the diversity in their scales and appearances,
we introduce a Multi-Scale Local Spatial Pathway. This pathway extracts multi-scale local features from the intra-window branch using convolutional layers with varying kernel sizes. Small kernel sizes ($k_s$) capture fine details of thin rain streaks, while larger kernels model broader and more diffuse rain patterns. This design enhances the model's ability to handle diverse rain scenarios while preserving local structural details.
\vspace{-0.1cm}
\begin{equation}
\begin{aligned}
 f_i = & \ \sigma(\Phi_{1d}^{k_i}(f_{i-1})), \quad i = 1, 2, \dots, n \quad \\
 f_0 = & \ \Phi_{1d}^{k_0}(\mathbb{L}(\text{RA}(f_\text{I}))) \quad     
% Y = \sigma  (\sum_{i=1}^{n} f_i ) 
f_{\text{ms}} = \sigma  (\sum\nolimits_{i=1}^{n} f_i )
\end{aligned}
\end{equation}

% \todo{
% \begin{equation}
% f_{\text{ms}} = \sigma \sum_{i} (\Phi_{1d}^{k_i} (\mathbb{L}(\text{RA}(f_\text{I})))) .
% \end{equation}
% }
The multi-scale features $f_{\text{ms}} $ are aggregated and fused with the global outputs $f_{\text{dual}}$ via residual addition, seamlessly integrating local details and global context. Finally, a linear layer ($\mathbb{L}$) refines the fused features, formulated as $f_{\text{out}} =  \mathbb{L}(f_{\text{ms}}+f_{\text{dual}})$, enhancing spatial precision while maintaining structural integrity.
% \begin{equation}
% f_{\text{out}} = Linear(f_{ms}+f_{dual})
% \end{equation}
% A dedicated pathway enhances fine-grained details by processing single-scan features through a lightweight Convolutional Neural Network (CNN)
% This local refinement pathway ensures that fine-grained details are preserved and enhanced, contributing to a more accurate and detailed deraining result.

\subsection{Training Loss}
\label{sec:loss}
\begin{figure}[htbp]
	\centering
    % \vspace{-0.4cm}
	\includegraphics[width=1.0\linewidth]{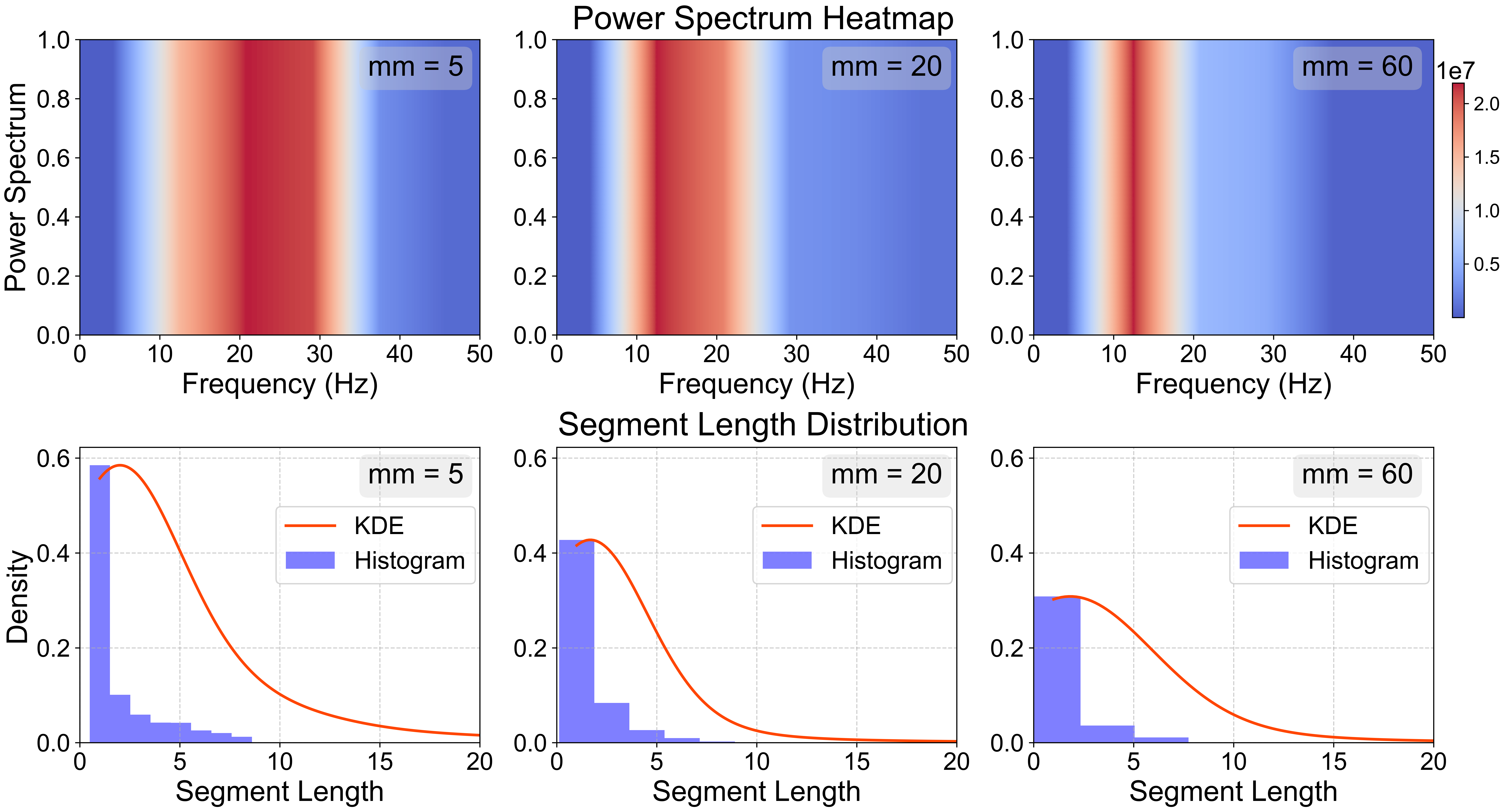}
    \vspace{-0.6cm}
      \caption{\textbf{Power Spectrum Heatmap and Segment Length Distribution for Rainfall Intensities (5, 20, 60 mm).} Top: Heatmaps show higher frequencies at peak power for lower densities, reflecting sparse rain. Bottom: Distributions of continuous signals reveal longer segments at higher densities, indicating sustained rain.
      % The x-axis represents segment length, and the y-axis represents normalized density.
      }
    % \vspace{-0.6cm}
	\label{fig:fft}
\end{figure}

 % Before performing the loss calculation, we first deserialize the obtained sequence labels to correspond to the initial event cloud. 
 Our training objective combines a standard binary cross-entropy loss ($\mathcal{L}_{\text{ce}}$) with a frequency-domain regularization term ($\mathcal{L}_{\text{fft}}$) to guide the model in learning both point-wise accuracy and global frequency patterns, forming the joint loss as $\mathcal{L} = \mathcal{L}_{\text{ce}} + \lambda \mathcal{L}_{\text{fft}}  $.

\textbf{Frequency-Domain Regularization:}
Existing image deraining methods integrate frequency layers to capture frequency-domain features~\cite{liu2020wavelet, zhen2024freqmamba,li2024fouriermamba}. However, for event camera with millions of event data, embedding Fast Fourier Transform (FFT) layer directly introduces significant computational overhead. Alternatively, we propose incorporating $\mathcal{L}_{\text{fft}}$ as a regularization term during model optimization, maintaining efficiency while mitigating overfitting.

While $\mathcal{L}_{\text{ce}}$ focuses on per-event accuracy, it may overlook global distribution patterns, resulting in unrealistic predictions like isolated rain events. In contrast, $\mathcal{L}_{\text{fft}}$ enforces statistical consistency by aligning frequency-domain distributions via amplitude and phase constraints:
\begin{equation}
    \mathcal{L}_{\text{fft}} = \frac{1}{L} \sum\nolimits_{i=1}^{L} \left( \frac{|\mathcal{F}(P)_i - \mathcal{F}(Y)_i|}{\max(|\mathcal{F}(P) - \mathcal{F}(Y)|, \epsilon)} + \epsilon' \right)^2,
\end{equation}
where $\mathcal{F}(\cdot)$ denotes the 1D FFT, $L$ is the sequence length, and $\epsilon$, $\epsilon'$ ensure numerical stability, and $Y$ and $P$ denote the ground truth labels and predicted outputs, respectively. Unlike traditional $\mathcal{L}_{\text{fft}}$ used in image reconstruction~\cite{jiang2021focal,zhen2024freqmamba}, our formulation operates directly on event labels, leveraging event cameras' microsecond-level resolution to capture rain dynamics.  Rainfall intensity directly correlated to event density and label $1$'s density (rain events labeled as $1$). As illustrated in Figure~\ref{fig:fft}, high-density rainfall manifests as low-frequency continuous patterns in the label sequence, while low-density rain exhibits high-frequency sparse patterns. This spectral distinction enables $\mathcal{L}_{\text{fft}}$ to align predictions with physically consistent global patterns~\cite{luo2025smartspr}.

\section{Dataset}
\label{sec:dataset}

\begin{figure}[t]
  \centering
% \fbox{\rule{0pt}{3in} \rule{0.9\linewidth}{0pt}}
   \includegraphics[width=1.0\linewidth]{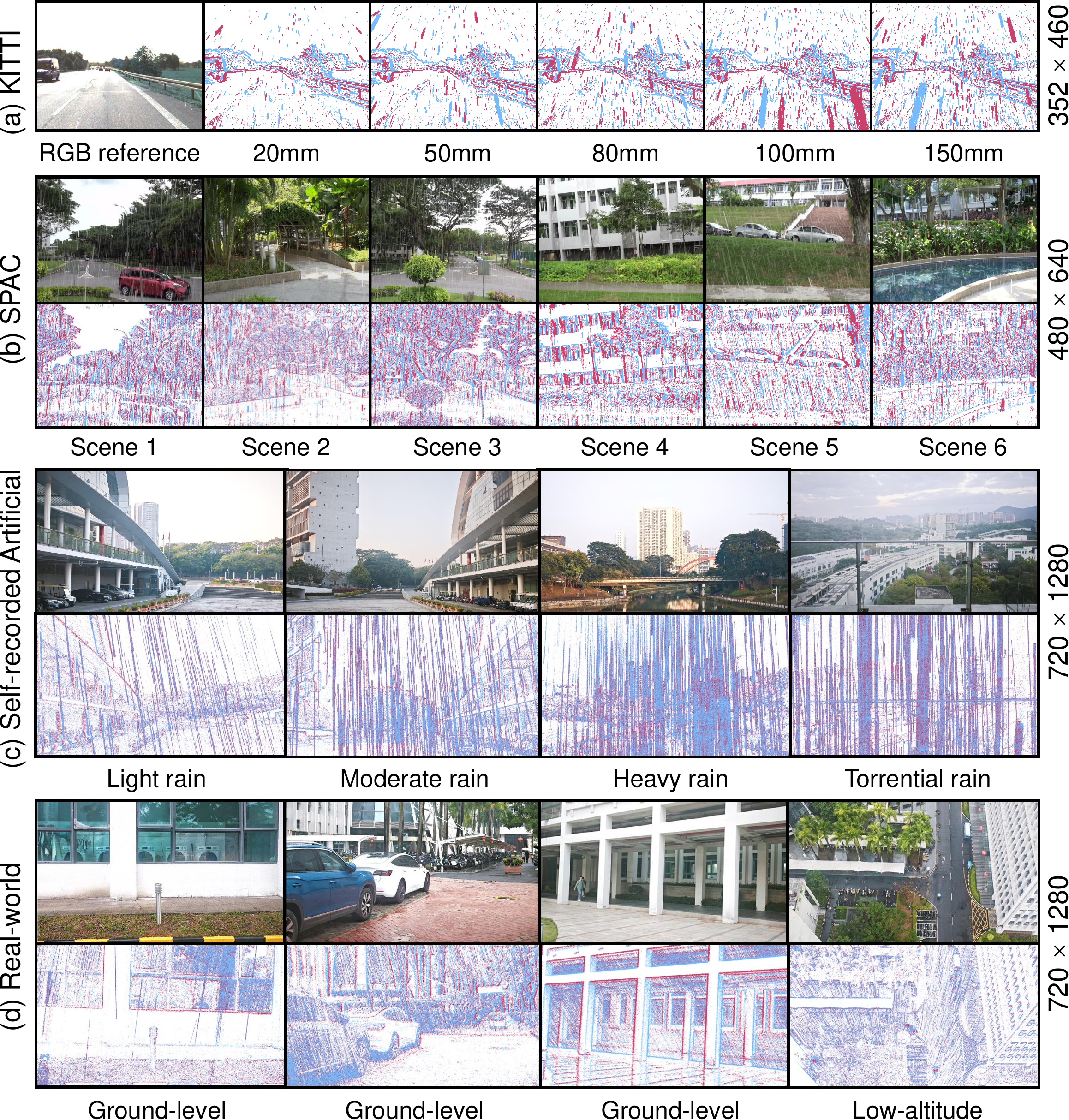}
   \vspace{-0.6cm}
  \caption{\textbf{EventRain-\datasetall: The First Point-based Event Deraining Dataset.} Additional dataset visualizations with varying rainfall intensities and scenarios are included in the Appendix.}
  % \caption{EventRain-8K: A Novel Point-based Event Deraining Benchmark} 或者叫这个
\label{fig:dataset}
\vspace{-0.6cm}
\end{figure}

% \begin{figure}[t]
%   \centering
% % \fbox{\rule{0pt}{3in} \rule{0.9\linewidth}{0pt}}
%    \includegraphics[width=1.0\linewidth]{ICCV2025-Author-Kit-Feb/img/dataset_v4.png}
%    \vspace{-0.3cm}
%   \caption{\textbf{1}}
%   % \caption{EventRain-8K: A Novel Point-based Event Deraining Benchmark} 或者叫这个
% \label{fig:dataset}
% \vspace{-0.6cm}
% \end{figure}

To support model training and testing, we published EventRain-\datasetall, the first point-based benchmark dataset for event-based deraining. The dataset comprises over \datasetsyn \  self-generated synthetic rain samples, over \datasetart \  self-recorded artificial rain samples, and over \datasetreal \  real rain samples.

% \subsection{Synthetic Dataset (labeled)}
\noindent\textbf{Synthetic Dataset (labeled).}
% Supervised event deraining relies on labeled data; however, annotating discrete, sparse, and highly dynamic event points is non-trivial. To address this, we employ a two-step approach to generate synthetic labeled data. We utilize the KITTI dataset~\cite{geiger2012we} and the SPAC dataset~\cite{chen2018robust} as clean video sources. Rain effects are synthesized using rain rendering simulators~\cite{tremblay2021rain, adobe2023aftereffects}. The KITTI dataset includes videos with varying camera motion speeds and rain intensities (\ref{fig:dataset}(a)), covering rainfall rates from light rain (5 mm/hour) to severe storms (200 mm/hour). The SPAC dataset features 16 diverse scenes, including cityscapes and natural environments, with each scene augmented by 3 to 4 distinct rain patterns.
% We then generate event sequences from these rainy videos using the Vid2E simulator~\cite{gehrig2020video}.
Supervised event deraining relies on labeled data; however, annotating discrete, sparse, and highly dynamic event points is non-trivial. To address this limitation, we use a two-step synthetic data generation approach. We first utilize the KITTI dataset~\cite{geiger2012we} and the SPAC dataset~\cite{chen2018robust} as clean video sources, then synthesize rain effects using state-of-the-art rendering simulators~\cite{tremblay2021rain, adobe2023aftereffects}. The KITTI rainy dataset provides videos with varying camera motion speeds and rain intensities (\ref{fig:dataset}(a)), covering rainfall rates ranging from light rain (5 mm/hour) to severe storms (200 mm/hour). The SPAC rainy dataset includes 16 diverse scenes, such as cityscapes and natural environments, each augmented with 3 to 4 distinct rain patterns (\ref{fig:dataset}(b)). Finally, we generate event sequences from these rainy videos using the Vid2E simulator~\cite{gehrig2020video}.
\begin{table*}[htbp]
% \footnotesize
\scriptsize
\renewcommand{\arraystretch}{1.0} % 行高
\setlength{\tabcolsep}{2.5pt} 
\begin{tabular*}{\textwidth}{@{\extracolsep{\fill}}c*{18}{c}@{}}
\toprule
\multirow{2}{*}{\centering Metrics} & \multicolumn{3}{c}{5mm} & \multicolumn{3}{c}{20mm} & \multicolumn{3}{c}{50mm} & \multicolumn{3}{c}{80mm} & \multicolumn{3}{c}{125mm} & \multicolumn{3}{c}{150mm} \\
\cmidrule(r){2-4}\cmidrule(l){5-7}\cmidrule(l){8-10}\cmidrule(l){11-13}\cmidrule(l){14-16}\cmidrule(l){17-19}
& SR $\uparrow$ & NR $\uparrow$ & DA $\uparrow$ & SR $\uparrow$ & NR $\uparrow$ & DA $\uparrow$ & SR $\uparrow$ & NR $\uparrow$ & DA $\uparrow$ & SR $\uparrow$ & NR $\uparrow$ & DA $\uparrow$ & SR $\uparrow$ & NR $\uparrow$ & DA $\uparrow$ & SR $\uparrow$ & NR $\uparrow$ & DA $\uparrow$ \\
\cmidrule{1-19}
% \multicolumn{1}{c}{Synthetic} & \multicolumn{3}{c}{5mm} & \multicolumn{3}{c}{20mm} & \multicolumn{3}{c}{50mm} & \multicolumn{3}{c}{80mm} & \multicolumn{3}{c}{125mm} & \multicolumn{3}{c}{150mm} \\
% \cmidrule(r){1-1}\cmidrule(l){2-4}\cmidrule(l){5-7}\cmidrule(l){8-10}\cmidrule(l){11-13}\cmidrule(l){14-16}\cmidrule(l){17-19}
% Metrics & SR $\uparrow$ & NR $\uparrow$ & DA $\uparrow$ & SR $\uparrow$ & NR $\uparrow$ & DA $\uparrow$ & SR $\uparrow$ & NR $\uparrow$ & DA $\uparrow$ & SR $\uparrow$ & NR $\uparrow$ & DA $\uparrow$ & SR $\uparrow$ & NR $\uparrow$ & DA $\uparrow$ & SR $\uparrow$ & NR $\uparrow$ & DA $\uparrow$ \\
% EvFlow\cite{Zhu-RSS-18} & \textbf{0.998} & 0.005 & 0.501 & 0.998 & 0.021 & 0.509 & 0.997 & 0.011 & 0.504 & 0.997 & 0.026 & 0.511 & 0.996 & 0.019 & 0.507 & 0.995 & 0.022 & 0.509  \\
TS\cite{6407468} & 0.888 & 0.265 & 0.576 & 0.887 & 0.305 & 0.596 & 0.883 & 0.231 & 0.557 & 0.881 & 0.271 & 0.576 & 0.877 & 0.237 & 0.556 & 0.872 & 0.243 & 0.557 \\
DWF\cite{9720086} & 0.703 & 0.394 & 0.549 & 0.734 & 0.464 & 0.599 & 0.755 & 0.352 & 0.553 & 0.777 & 0.432 & 0.604 & 0.786 & 0.371 & 0.578 & 0.782 & 0.375 & 0.578 \\
Knoise\cite{8244294} & 0.860 & 0.375 & 0.618 & 0.870 & 0.377 & 0.624 & 0.884 & 0.241 & 0.563 & 0.890 & 0.257 & 0.574 & 0.901 & 0.211 & 0.556 & 0.896 & 0.214 & 0.555\\
Ynoise\cite{Feng2020EventDB} & 0.679 & 0.533 & 0.606 & 0.677 & 0.568 & 0.623 & 0.663 & 0.481 & 0.572 & 0.658 & 0.525 & 0.591 & 0.647 & 0.482 & 0.564 & 0.634 & 0.487 & 0.561 \\
RED\cite{baldwin2020event} &  0.862 & 0.172 & 0.517 & 0.853 & 0.251 & 0.552 & 0.833 & 0.208 & 0.520 & 0.822 & 0.185 & 0.503 & 0.808 & 0.207 & 0.507 & 0.789 & 0.208 & 0.499  \\ 
\cmidrule{1-19}
EDnCNN\cite{baldwin2020event} & 0.968& \underline{0.905}& \underline{0.937}& 0.954&\underline{0.904} & \underline{0.929}&\underline{0.948} &\underline{0.888} &\underline{0.918} & \underline{0.935}& \underline{0.870}& \underline{0.903}& \textbf{0.933}& \underline{0.867}& \underline{0.900}&\textbf{0.929}& \underline{0.843}& \underline{0.886} \\
AEDNet\cite{fang2022aednet} & 0.941 & 0.850 & 0.891 & 0.938 & 0.876 & 0.907 & 0.928 & 0.732 & 0.830 & 0.925 & 0.681 & 0.803 & 0.922 & 0.624 & 0.773 & 0.923 & 0.547 & 0.735 \\ 
EDformer\cite{jiang2024edformer}  & \underline{0.981} & 0.818& 0.899&\underline{0.962} & 0.832 & 0.897& 0.924 & 0.844 & 0.884& 0.894& 0.834& 0.864&0.876 & 0.840& 0.858&0.839 & 0.834& 0.836\\   
Ours &\textbf{0.994} & \textbf{0.914} & \textbf{0.954} & \textbf{0.978} & \textbf{0.915} & \textbf{0.947} & \textbf{0.955} & \textbf{0.911} & \textbf{0.933} & \textbf{0.940} & \textbf{0.903} & \textbf{0.922} & \underline{0.918} & \textbf{0.898} & \textbf{0.908} & \underline{0.908} & \textbf{0.895} & \textbf{0.902}\\
\bottomrule
\end{tabular*}
% \caption{Performance comparison of different derain methods under different precipitation intensities.}
\vspace{-0.3cm}
\caption{\textbf{Quantitative Comparisons on Synthetic Dataset with Varying Rainfall Intensities.} We mark the \textbf{best} and \underline{second best} results. }
% \caption{The mean SR, NR and DA results of different event deraining methods on synthetic dataset with varying rain intensities. We mark the \textbf{best} and \underline{second best} results.}
\label{tab:accuracy}
\vspace{-0.6cm}
% \caption{THE MEAN SR, NR AND DA RESULTS OF DIFFERENT EVENT DERAINING METHODS ON SYNTHETIC DATASET WITH VARYING PRECIPITATION INTENSITIES.WE MARK THE \textbf{BEST} ANE \underline{SENCOND BEST}. }
\end{table*}

\noindent\textbf{Self-recorded Artificial Dataset (labeled).}
% \todo{While the synthetic dataset simulates diverse rain intensities, its event generation mechanism does not fully capture the spatial coherence and motion dynamics of real rain. }To bridge this gap, we constructed an artificial rainfall dataset using controlled experiments.
To bridge the domain gap between synthetic and real-world event data for model training, we introduce a self-recorded artificial dataset, capturing real event camera data under controlled artificial rainfall.
A showerhead was employed to simulate raindrop free-fall motion, captured by a Prophesee EVK4 event camera (1280 $\times$ 720 resolution, 10k fps, 120dB dynamic range) for both rainy and rain-free data.  Slight camera vibrations were applied during captures to obtain complete scene details.
Utilizing the K-Nearest Neighbors (KNN) algorithm, background events in rainy data were identified via spatiotemporal alignment with rain-free data, enabling background labeling and isolating rain events.
The dataset focuses on urban outdoor scenes (buildings, trees, cars, lakes) and covers diverse rainfall intensities across four scenarios (Figure \ref{fig:dataset}(c)).
\begin{figure}[t]
\centering
\vspace{0.2cm}
    \includegraphics[width=0.8\linewidth]{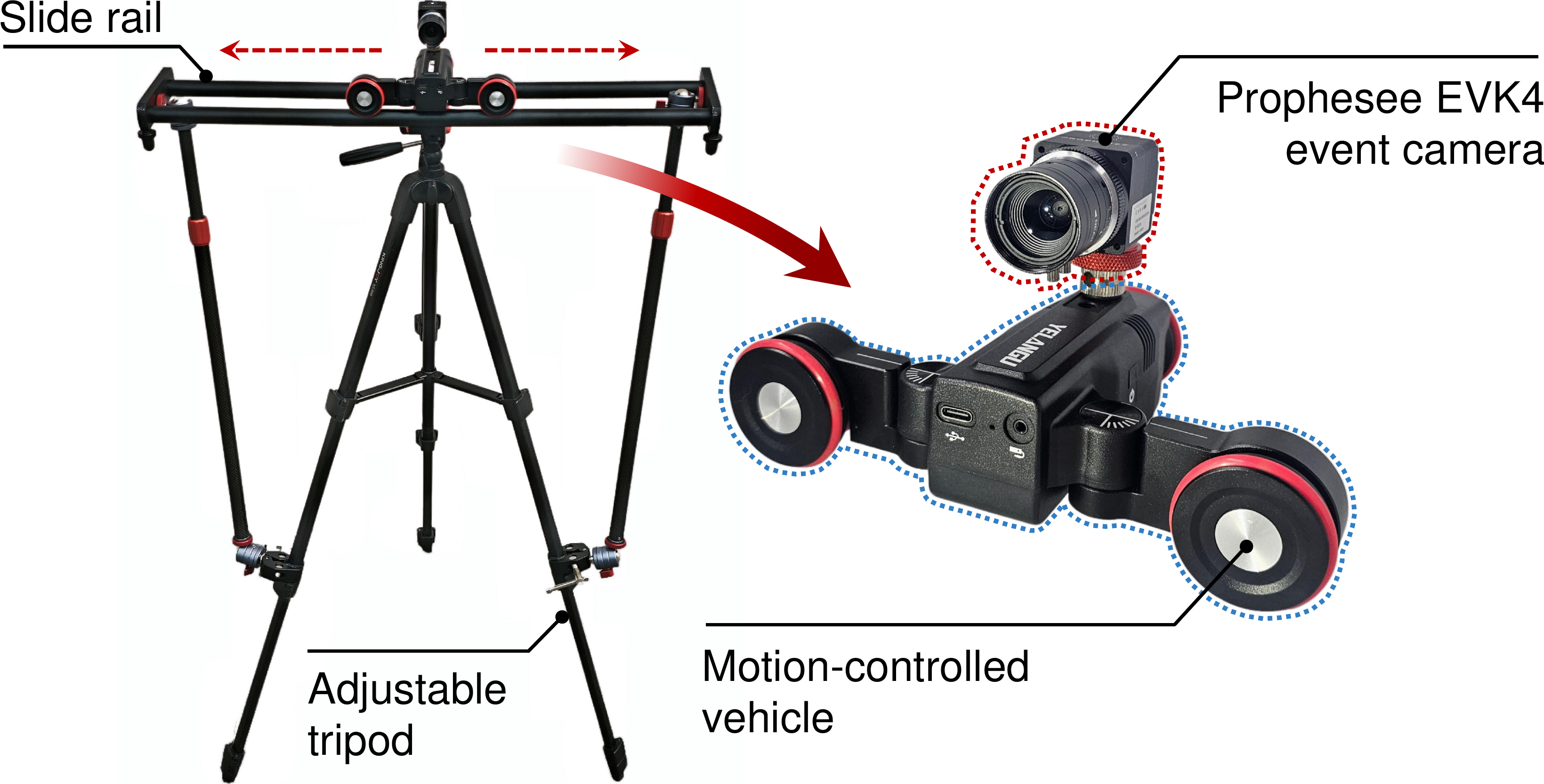}
\caption{Our fixed-point \& speed data acquisition system.}
\label{fig:tripod}
\vspace{-0.6cm}
\end{figure}
\noindent\textbf{Real-world Dataset (label-free).}
% Our real-world dataset aims to demonstrate the validity and generalization of \model \  for real rain. 
To validate event deraining effectiveness and generalization, we collect a real-world event dataset.
As illustrated in Figure \ref{fig:tripod}, 
% we developed a data acquisition system featuring fixed-point and fixed-speed (10 cm/s) cruising. 
% The Prophesee EVK4 event camera mounted on a motion-controlled vehicle traversed a slide rail, enabling the simultaneous capture of background and rain data under controlled dynamic conditions
we designed a data acquisition system with fixed-point and fixed-speed (10 cm/s) cruising. 
% A Prophesee EVK4 event camera mounted on a motion-controlled vehicle moved along a slide rail, capturing both background and rain data under controlled dynamic conditions. 
% A Prophesee EVK4 event camera, mounted on a motion-controlled vehicle, traversed a slide rail to simultaneously capture background and rain data under controlled dynamic conditions.
We mounted a Prophesee EVK4 event camera on a motion-controlled vehicle, which traversed a slide rail to capture background and rain data simultaneously under controlled dynamic conditions.
% with comparable density, reflecting dynamic application scenarios. 
% The dataset spans 26 ground-level and low-altitude scenes, including rainy event sequences across varying intensities (Figure \ref{fig:dataset}(c)).
The dataset includes diverse viewpoints and rain intensities, covering a wide range of real-world scenarios (Figure \ref{fig:dataset}(d)).
% to evaluate model performance and generalization. 
Additionally, it incorporates \datasetsnow \  samples of real snow data captured to evaluate the model's generalization capability.

% Representative samples are visualized in Figure \ref{fig:dataset}.

% As shown in Figure \textcolor{red}{\ref{fig:tripod}}, we constructed a data acquisition system for fixed-point and fixed-speed cruising, where the event camera was mounted on a motion-controlled vehicle and traversed along a slide rail. Through the translational movement of the camera, we simultaneously captured background and rain data with comparable density, aligning with the dynamic application scenarios of event cameras.

% This dataset encompasses both ground-level and low-altitude scenes, and includes collections of rain and snow data across various rainfall intensities. These data serve to evaluate the performance and demonstrate the generalization capability of the model. We present visualizations of representative real rainfall samples in Figure \textcolor{red}{AA}.

% \input{ICCV2025-Author-Kit-Feb/tab/0_accuracy}
% \input{ICCV2025-Author-Kit-Feb/tab/1_efficiency}
% \input{ICCV2025-Author-Kit-Feb/tab/2_ablation}
\vspace{-0.3cm}
\section{Experiments}
\label{sec:experiment}

\subsection{Experiments Setups}
\noindent \textbf{Evaluation Metrics.} We use event Denoising Accuracy (DA), Signal Retention (SR) and Noise Removal (NR) to measure the event rain removal performance on \model. There is $ \text{DA} = \frac{1}{2}(\text{SR} + \text{NR}) = (\frac{\text{PB}}{\text{TB}} + \frac{\text{PR}}{\text{TR}})$, where $ \text{PB},\text{TB},\text{PR},\text{TR} $  are the predicted background, the predicted rain, the real background, and the number of real rain, respectively. 
DA measures rain removal while preserving real motion (e.g., vehicles, pedestrians) and structural details (e.g., roads, building profiles). SR and NR analyze signal retention and noise suppression, respectively.
% The core metric DA is used to measure whether the algorithm can remove rain while maintaining the integrity of real motion information (such as vehicles, pedestrians) and structural details (such as roads, building profiles) in the scene, while SR and NR provide fine-grained analysis from the perspective of signal retention and noise suppression, respectively. 
% We evaluated the rain removal effect of the model under different rainfall and different scenarios. Moreover, the visualization results were qualitatively evaluated on real rainfall datasets. / put it in Quantitative Results

\noindent \textbf{Implementation Details.}
% Our method is implemented in PyTorch and trained with Adam optimizer\tocite, 
% In our experiments, the total batch size is set to 6. The entire training process spans 200 epochs, with evaluations conducted every 20 epochs. 
% Our network is implemented in PyTorch and trained using the AdamW optimizer with an initial learning rate of 0.00048 and a weight decay coefficient of $5 \times 10 ^{-3} $. 
Our network is implemented in PyTorch and trained using the AdamW optimizer with an initial learning rate of $4.8 \times 10 ^{-4}$ and a weight decay coefficient of $5 \times 10 ^{-3} $. 
% The learning rate is scheduled using OneCycleLR, with a 5$\%$ warm-up phase and cosine annealing for adjustment. 
The entire training is conducted on six NVIDIA RTX A6000 GPUs for 50 epochs with a batch size of 6.
 The model processes a sequence of five temporal windows as input,  with each window spanning 0.1 seconds. In the multi-spatial-scale module of the MS3M, we employ kernel sizes of (1, 3, 5) to capture features at different spatial resolutions.
 % The architecture of \model \ remains consistent with the U-Net\tocite framework.
 Network details can be found in the Appendix.
%
% It consists of two stage encoders and one decoders, with respective block depths of [2, 4] and [2]. 
% For the loss function, we select both pixel-wise cross-entropy loss and Lovasz loss for multi-class tasks.
% We implement all experiments using the PyTorch framework, running on 6 servers equipped with NVIDIA RTX A6000 GPUs.
\subsection{Comparisons with State-of-The-Art Methods}
\noindent \textbf{Baselines.}
As this work represents the first attempt at point-based event deraining, there are no directly comparable baselines in the literature. Alternatively, we adapt state-of-the-art event denoising methods, as deraining can be viewed as a specialized case of noise removal. These include filter-based methods TS~\cite{6407468}, DWF~\cite{9720086}, Knoise~\cite{8244294}, Ynoise~\cite{Feng2020EventDB}, RED~\cite{baldwin2020event} and learning-based algorithms EDnCNN~\cite{baldwin2020event}, AEDNet~\cite{fang2022aednet}, EDformer~\cite{jiang2024edformer}. 
% Additionally, we qualitatively compare with DistillNet~\cite{ruan2024distill}, a voxel-based deraining method using GANs. However, its voxelized event processing paradigm prevents quantitative evaluation using event point denoising metrics.
Additionally, we qualitatively compare with DistillNet~\cite{ruan2024distill}, a voxel-based GAN deraining method, but its voxelized processing is incompatible with event point denoising metrics, preventing quantitative evaluation.

\noindent \textbf{Results on Synthetic and Artificial Datasets.}
\begin{figure*}[t]
\vspace{-0.3cm}
  \centering
   \begin{overpic}[width=1.0\linewidth]{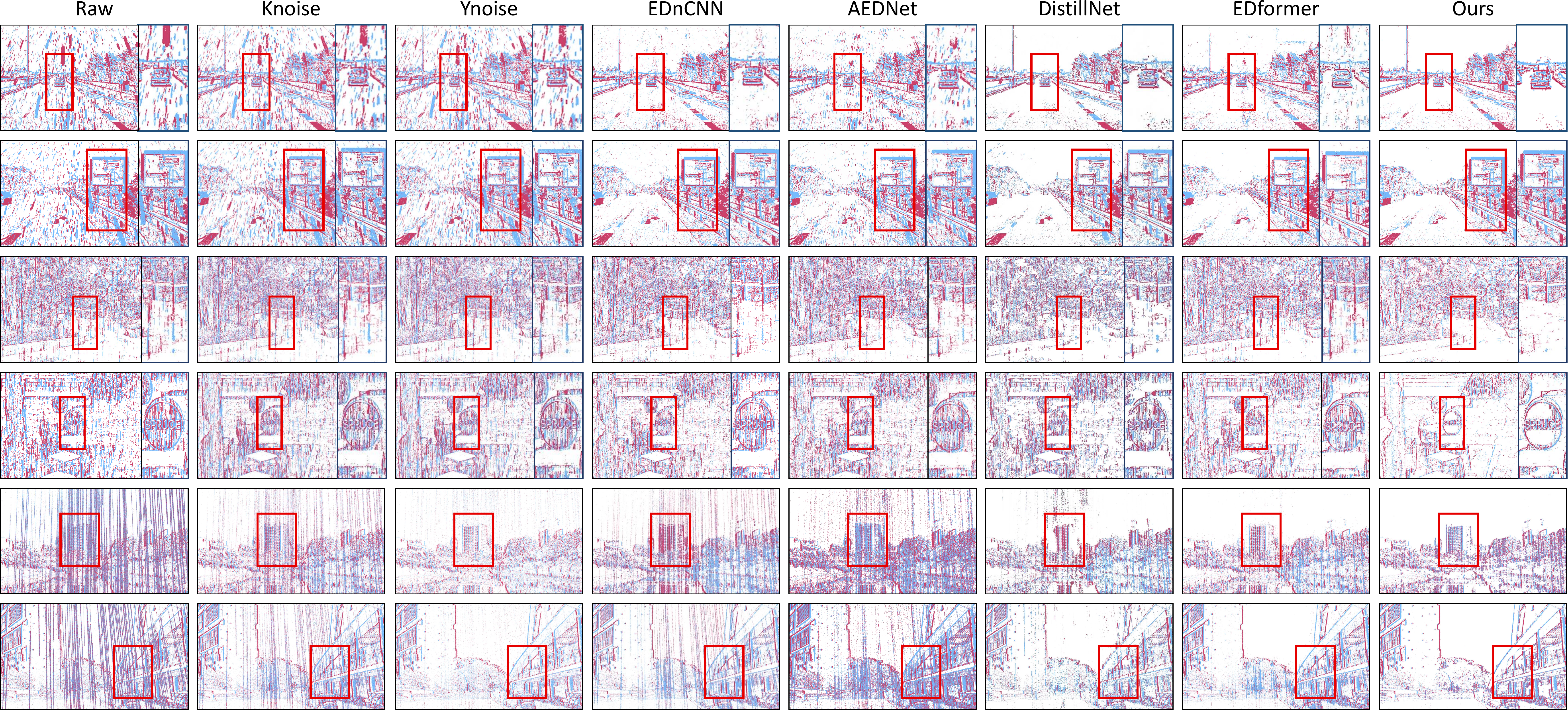}
    \put(20.7,44.5){\color{red}\footnotesize\textbf{\cite{8244294}}}
    \put(33.4,44.5){\color{red}\footnotesize\textbf{\cite{Feng2020EventDB}}}
    \put(46.5,44.5){\color{red}\footnotesize\textbf{\cite{baldwin2020event}}}
    \put(58.7,44.5){\color{red}\footnotesize\textbf{\cite{fang2022aednet}}}
    \put(71.6,44.5){\color{red}\footnotesize\textbf{\cite{ruan2024distill}}}
    \put(84.3,44.5){\color{red}\footnotesize\textbf{\cite{jiang2024edformer}}}
\end{overpic}
\vspace{-0.7cm}
  \caption{\textbf{Qualitative Comparison on Synthetic and Artificial Datasets.} \model\ demonstrates superior deraining capabilities across diverse structural contours and complex textures with enhanced detail preservation. Please zoom in for details.}
  \vspace{-0.5cm}
\label{fig:synthetic}
\end{figure*}
In Table~\ref{tab:accuracy}, we present SR, NR, and DA results on the synthetic dataset with six rain intensities, ranging from light rain to heavy rain.  Our method achieves state-of-the-art performance, with an average SR/NR/DA of 0.95/0.91/0.93. It is notable that learning-based algorithms outperform filtering methods due to their ability to leverage supervised training. When training EDnCNN~\cite{baldwin2020event},  we replaced its original event probability mask (EPM) labeling framework with our ground truth labels, which improved its performance. In addition, we provide qualitative results in Figure~\ref{fig:synthetic}. 
Filtering methods are ineffective for rain removal, as rain streaks exhibit distinct spatiotemporal patterns that deviate from typical noise. 
% Traditional filtering methods are ineffective for rain removal, as they uniformly filter both rain and background, failing to address the distinct spatiotemporal patterns of rain streaks that differ from typical noise.
Other learning-based methods yield incomplete deraining results, often retaining residual rain streaks. In contrast, our method effectively removes rain streaks and restores details through multi-scale temporal and spatial modeling combined with motion and appearance perception.

\begin{figure*}[htbp]
  \centering
   % \includegraphics[width=1.0\linewidth]{ICCV2025-Author-Kit-Feb/img/results_v5.png}
   % \begin{overpic}[width=1.0\linewidth]{ICCV2025-Author-Kit-Feb/img/vis_real_v3.png}
      \begin{overpic}[width=1.0\linewidth]{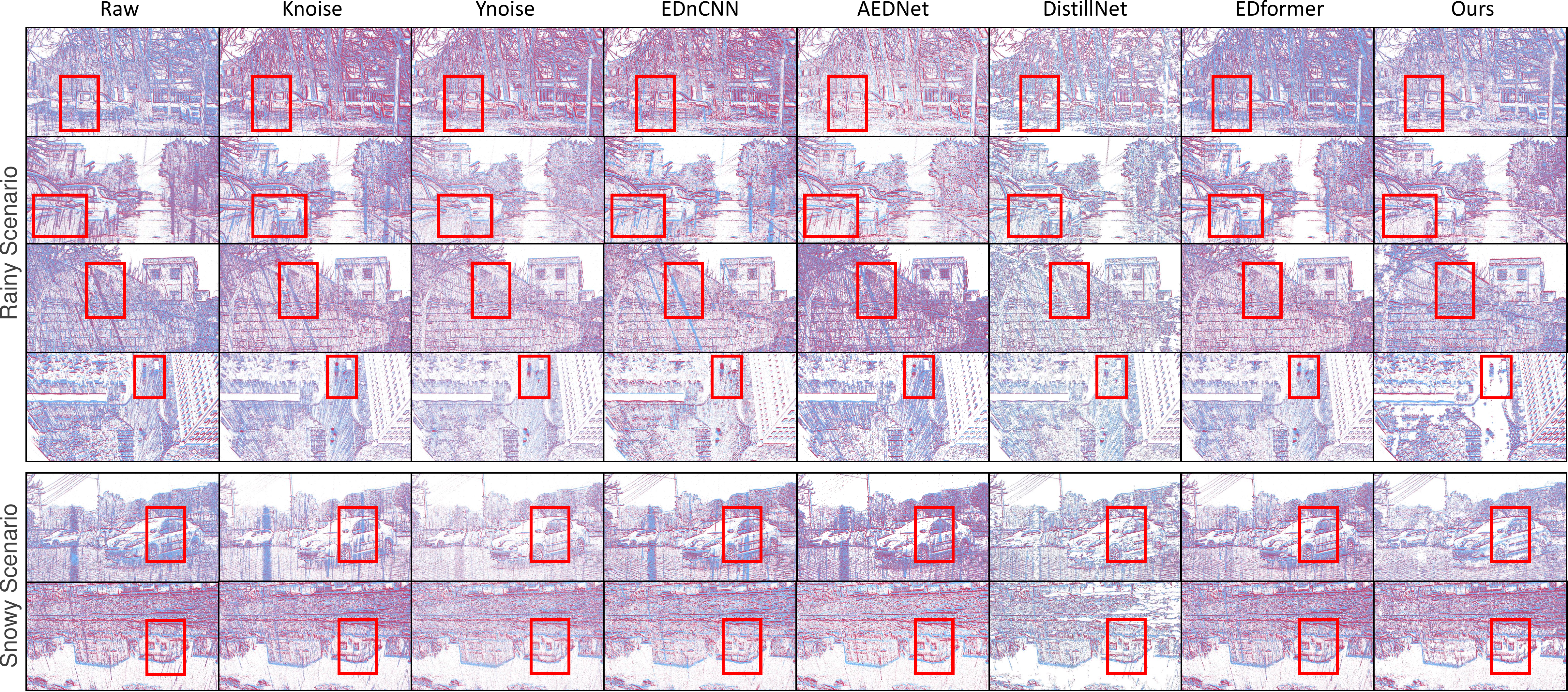}
    \put(22.5,43.2){\color{red}\footnotesize\textbf{\cite{8244294}}}
    \put(34.7,43.2){\color{red}\footnotesize\textbf{\cite{Feng2020EventDB}}}
    \put(47.7,43.2){\color{red}\footnotesize\textbf{\cite{baldwin2020event}}}
    \put(59.7,43.2){\color{red}\footnotesize\textbf{\cite{fang2022aednet}}}
    \put(72.3,43.2){\color{red}\footnotesize\textbf{\cite{ruan2024distill}}}
    \put(84.7,43.2){\color{red}\footnotesize\textbf{\cite{jiang2024edformer}}}
\end{overpic}
  % \caption{Qualitative results on real rain and snow datasets. Our method gets the best deraining and desnowing results. Please zoom in for details.}
  \vspace{-0.5cm}
  \caption{\textbf{Qualitative Comparison on Real-World Datasets.} The top four rows highlight our method's superior deraining performance on real rain datasets, while the last two rows showcase its generalization to snow scenes. Please zoom in for details.}
  \vspace{-0.5cm}
\label{fig:realworld}
\end{figure*}

\noindent \textbf{Generalization on Real-World Dataset.}
To validate practical applicability, we compare our method against baselines on the real-world data of EventRain-\datasetall. Figure~\ref{fig:realworld} (top four rows) isualizes results for four scenarios, including three ground-level and one low-altitude viewpoints. Our method achieves superior performance compared to competitors. In the low-altitude scenario, our method effectively eliminates rain streaks while maintaining the structural integrity of static objects (e.g., buildings) and preserving the motion patterns of dynamic objects (e.g., vehicles). 
Filter-based methods treat both background and rain equally, resulting in effects similar to downsampling. While EDnCNN~\cite{baldwin2020event}, AEDNet~\cite{fang2022aednet}, and EDformer~\cite{jiang2024edformer} are trained on synthetic data, their denoising-focused designs fail to generalize well to real rain, further hindered by the synthetic-to-real domain gap. DistillNet~\cite{ruan2024distill} removes large rain streaks via voxel-based GANs but loses background details and introduces artifacts.  In contrast, our method, designed for point-based event deraining, effectively removes rain while preserving both background details and dynamic motion content, demonstrating robust real-world performance. 
% \begin{table*}[htbp]
% \small
% \renewcommand{\arraystretch}{1}
% \begin{tabularx}{\linewidth}{@{}c*{6}{>{\centering\arraybackslash}X}|*{4}{>{\centering\arraybackslash}X}@{}}
% \toprule
% & EvFlow & TS &  DWF & KNoise & YNoise & RED & EDnCNN &  AEDNet& EDformer & Ours \\
% % \cmidrule[0.5pt](r){1-7}\cmidrule[0.5pt](l){8-10}
% \midrule
% \textbf{GFLOPs} & --- & --- & --- & --- & --- & --- &   & 1.12& & 45.72 \\
% \textbf{Parameters} & --- & --- & --- & --- & --- & --- &   & 45.87M&  & 264632\\
% \textbf{Inference time} & 10.5957 & 0.1296 & 0.0954 & 0.0198 & 0.0513 & 2.2716 &  & 283.492 & & 1.03218 \\
% \bottomrule
% \end{tabularx}
% \caption{Comparison of different denoising methods. We mark the \textbf{best} and \underline{second best} results.}
% \label{tab:comparison}
% \end{table*}

\begin{table}[t]
% \footnotesize
\scriptsize
\renewcommand{\arraystretch}{1.0}
\begin{tabular}{>{\centering\arraybackslash}>{\centering\arraybackslash}p{0.15\linewidth}>{\centering\arraybackslash}p{0.08\linewidth}>{\centering\arraybackslash}p{0.11\linewidth}>{\centering\arraybackslash}p{0.22\linewidth}>{\centering\arraybackslash}p{0.18\linewidth}}
\toprule
& \textbf{GFLOPs} & \textbf{Parameters} & \textbf{Inference time (s)} & \textbf{Relative speed}\\
\midrule
% EvFlow\cite{Zhu-RSS-18} & N/A & N/A & 10.5957 \\
TS\cite{6407468} & N/A & N/A & 0.1296 & 1.0 $\times$\\
DWF\cite{9720086} & N/A & N/A & 0.0954 & 1.36$\times$\\
KNoise\cite{8244294} & N/A & N/A & \textbf{0.0198} & \textbf{6.55}$\times$\\
YNoise\cite{Feng2020EventDB} & N/A & N/A & 0.0513 & \underline{2.53}$\times$\\
RED\cite{baldwin2020event} & N/A & N/A & 2.2716 & 0.06$\times$\\
\midrule
EDnCNN\cite{baldwin2020event} & 234.51&614.55K & 20.1885 & 1.0$\times$\\
AEDNet\cite{fang2022aednet} & 4400.46 & 45.87M & 43.4250 & 0.46$\times$\\
DistillNet\cite{ruan2024distill} & 255.17 & 18.96M & \underline{0.2029} & \underline{99.50$\times$}\\
EDformer\cite{jiang2024edformer} & \underline{8.41}& \textbf{49.80K}& 2.4943 & 8.09$\times$\\
Ours & \textbf{6.23} & \underline{264.63K} & \textbf{0.0987} & \textbf{204.54$\times$}\\
\bottomrule
\end{tabular}
\vspace{-0.3cm}
% \caption{Comparison of different denoising methods. We mark the \textbf{best} and \underline{second best} results.}
\caption{Model complexity comparisons with previous model. }
\label{tab:efficiency}
\vspace{-0.6cm}
\end{table}

\noindent \textbf{Model Complexity and Efficiency Comparison.}
In Table \ref{tab:efficiency}, we compare the computational efficiency of our method and baselines in terms of FLOPs, parameters, and inference time for processing 100K events on an NVIDIA RTX A6000 GPU. Filtering-based methods
% , such as TS~\cite{6407468}, DWF~\cite{9720086}, KNoise~\cite{8244294}, YNoise~\cite{Feng2020EventDB}, and RED~\cite{baldwin2020event}, 
enable fast inference
but suffer from poor deraining performance due to lack of task-specific optimization. EDnCNN~\cite{baldwin2020event} and AEDNet~\cite{fang2022aednet} incur high computational overhead from encoding or sampling spatial-temporal neighborhoods, resulting in high FLOPs and slow inference. Transformer-based EDformer~\cite{jiang2024edformer} reduces FLOPs and parameters but suffers from high latency due to the quadratic complexity of its attention mechanism. DistillNet~\cite{ruan2024distill} achieves a lower inference time by voxelizing events but at the cost of high computational complexity, large model size, and loss of fine-grained temporal precision. In contrast, our method achieves SOTA performance with only 6.23G FLOPs and 0.26M parameters, processing 100K events in 0.0987s and scaling to 1M events in 0.398s. 
Compared to EDnCNN and AEDNet,
% our method reduces FLOPs by 37.6$\times$ and 706.3$\times$, 
our method reduces FLOPs to 2.66\% and 0.14\% of theirs, respectively, while achieving 204.5$\times$ and 440.0$\times$ faster inference.
% our method achieves only 2.66\% and 0.14\% of their FLOPs, respectively,
% and speeds up inference by 204.5$\times$ and 440.0$\times$. 
Compared to the lightweight EDformer, our method achieves a 25.3$\times$  speedup with comparable model size, 
% Even against the lightweight EDformer, our method achieves a 25.3$\times$ speedup with comparable model size, 
excelling in both accuracy and computational efficiency.
% Even compared to the lightweight EDformer, our method achieves 25.3× faster inference with 74\% fewer parameters. Furthermore, our method outperforms filtering-based approaches in deraining performance while maintaining competitive efficiency, demonstrating a superior balance between accuracy and computational cost. 

\vspace{-0.2cm}
\subsection{Ablation Studies}
% \begin{table}[ht]
% \centering
% \begin{tabular}{lccc|ccc}
% \toprule
% Model & GMB & LMB & DCL & PSNR$\uparrow$ & SSIM$\uparrow$ & LPIPS$\downarrow$ \\
% \midrule
% M1    &  &  &  & 29.35 & 0.9118 & 0.0893 \\
% M2    & \checkmark &  &  & 30.86 & 0.9277 & 0.0774 \\
% M3    & \checkmark & \checkmark &  & 31.04 & 0.9302 & 0.0753 \\
% M4    & \checkmark & \checkmark & \checkmark & 31.79 & 0.9361 & 0.0704 \\
% \textbf{Ours} & \checkmark & \checkmark & \checkmark & 32.04 & 0.9376 & 0.0684 \\
% \bottomrule
% \end{tabular}
% \caption{Performance comparison of different model variants.}
% \end{table}
% \begin{table}[h]
%     \centering
%     \begin{tabular}{l|ccc|ccc}
%         \toprule
%         Model & GMB & LMB & DCL & PSNR$\uparrow$ & SSIM$\uparrow$ & LPIPS$\downarrow$ \\
%         \midrule
%         M1    &     &     &     & 29.35 & 0.9118 & 0.0893 \\
%         M2    & \checkmark  &     &     & 30.86 & 0.9277 & 0.0774 \\
%         M3    &     &\checkmark  &\checkmark   & 31.04 & 0.9302 & 0.0753 \\
%         M4    & \checkmark   & \checkmark   &     & 31.79 & 0.9361 & 0.0704 \\
%         Ours  & \checkmark   & \checkmark  & \checkmark   & 32.04 & 0.9376 & 0.0684 \\
%         \bottomrule
%     \end{tabular}
% \end{table}
\begin{table}[t]
\scriptsize
    \centering
     \renewcommand{\arraystretch}{1.0}
     \resizebox{\linewidth}{!}{
    \begin{tabular}{l|ccc|ccc}
        \hline
        Model & STDF & MS3M & $\mathcal{L}_{\text{fft}}$ & SR$\uparrow$ & NR$\uparrow$ & DA$\uparrow$ \\
        \hline
        M1    &     &     &     & 0.9123 & 0.8394 & 0.8759 \\
        M2    & \checkmark &     &     & 0.8983 & 0.8747 & 0.8865 \\
        M3    &     & \checkmark &  & 0.8941 & 0.8906 & 0.8923 \\
        M4    & \checkmark & \checkmark &     & 0.9046 & 0.8954 & 0.9000 \\
                \hline
        Ours  & \checkmark & \checkmark & \checkmark & 0.9080 & 0.8949 & 0.9015 \\
        \hline
    \end{tabular}
    }
    \vspace{-0.3cm}
    \caption{Ablation about each module in \model \  on the EventRain-\datasetall \  validation set.}
    \label{tab:comparison}
    \vspace{-0.7cm}
\end{table}
\noindent \textbf{Baseline  Design.} 
% We conduct ablation studies to evaluate the effectiveness of three key components in our network: the Spatio-Temporal Decoupling and Fusion (STDF) module, the Multi-Scale State Space Model (MS3M), and the Frequency Loss.
% We first establish a baseline (``M1'') by removing all three components. Next, we introduce the STDF module to create ``M2''. Then, ``M3'' is constructed by integrating MS3M into ``M1". Subsequently, ``M4" is built by adding MS3M to ``M3". Finally, we incorporate the Frequency Loss into ``M4" to complete our full model.
% Specifically, without STDF, temporal and spatial features are directly summed. Without MS3M, it is replaced with the original MSSM block~\cite{zeng2024mambamos}. Without frequency loss, the training loss is reduced to cross-entropy loss alone.
We conduct systematic ablation studies to validate the efficacy of three proposed components: the Spatio-Temporal Decoupling and Fusion (STDF) module, Multi-Scale State Space Model (MS3M), and Frequency Loss ($\mathcal{L}_{\text{fft}}$). Our baseline model (``M1") removes all three components, employing naive summation for spatio-temporal feature fusion, reverting to the original MSSM block~\cite{zeng2024mambamos} for temporal modeling, and using only cross-entropy loss. Building upon ``M1", ``M2" introduces the STDF module to embed structured spatio-temporal features, while ``M3" integrates the MS3M architecture for enhanced multi-scale modeling. And ``M4" combines both STDF and MS3M components. Finally, we incorporate the Frequency Loss into ``M4" to complete our full model. 

\noindent \textbf{Quantitative Comparison.} 
% From Table \ref{tab:comparison}, we observe that applying only STDF or MS3M improves performance over the baseline by 1.21\% and 1.87\% in DA, respectively, demonstrating their effectiveness. ``\textbf{M3}'' achieves further improvement, indicating that MS3M effectively captures deep-level spatio-temporal information and scales across diverse rain intensities and appearances. ``\textbf{M4}" achieves better results than ``\textbf{M3}", showcasing the benefits of combining STDF with MS3M for modeling multi-scale rain and event characteristics. Finally, our full network surpasses ``\textbf{M4}", underscoring the effectiveness of the frequency-domain regularization term in guiding the model through both amplitude and phase constraints.
From Table \ref{tab:comparison}, we observe that both STDF (``M2”) and MS3M (``M3”) improve the baseline performance by 1.21\% and 1.87\% in DA, respectively, validating their individual contributions. ``M3'' achieves greater improvement, demonstrating MS3M's ability to capture deep spatio-temporal features and adapt to diverse rain patterns. By integrating STDF with MS3M, ``M4" further enhances performance, demonstrating the complementary nature of these components in modeling multi-scale rain and event characteristics. Finally, our full network outperforms ``M4", which verifies the importance of the frequency-domain regularization term in leveraging both amplitude and phase constraints for improved learning.

\begin{table}[t]
% \footnotesize
\scriptsize
\renewcommand{\arraystretch}{1.0}
\begin{tabular}{>{\centering\arraybackslash}>{\centering\arraybackslash}p{0.26\linewidth}>{\centering\arraybackslash}p{0.18\linewidth}>{\centering\arraybackslash}p{0.18\linewidth}>{\centering\arraybackslash}p{0.18\linewidth}}
\toprule
 \textbf{Window Number} & \textbf{Num = 3} & \textbf{Num = 5} &\textbf{Num = 8} \\
\midrule
% EvFlow\cite{Zhu-RSS-18} & N/A & N/A & 10.5957 \\
DA & 0.8268 & 0.9015 & 0.9330 \\
Improve (\%)& - & 9.03 & 12.84 \\
Inference time (ms) & 274.02 & 394.16 & 525.68 \\
Relative speed  & - & 0.70 $\times$ & 0.52 $\times$ \\
\bottomrule
\end{tabular}
\vspace{-0.3cm}
% \caption{Comparison of different denoising methods. We mark the \textbf{best} and \underline{second best} results.}
\caption{Ablation about time window number on validation set.}
\label{tab:ablation2}
\vspace{-0.75cm}
\end{table}

% \begin{table}[t]
% % \footnotesize
% \scriptsize
% \renewcommand{\arraystretch}{1.0}
% \begin{tabular}{>{\centering\arraybackslash}>
% {\centering\arraybackslash}p{0.1\linewidth}>{\centering\arraybackslash}p{0.15\linewidth}>{\centering\arraybackslash}p{0.15\linewidth}>
% {\centering\arraybackslash}p{0.2\linewidth}>{\centering\arraybackslash}p{0.2\linewidth}
% }
% \toprule
%  \textbf{Window Num} & DA & Improve(\%) & Inference time(ms) & Speed Decrease(\%) \\
% \midrule

% \textbf{3} & 0.8268 & - & 274.02 & - \\% 0.9015 & 0.9330 \\
% \textbf{5}& 0.9015 & 9.03 & 394.16 & 30.49 \\ %- & 9.03 & 12.84 \\
% \textbf{8}& 0.9930 & 12.84 & 525.68 & 47.87 \\ %274.02 & 394.16 & 525.68 \\
% % Speed Decrease(\%)  & - & 30.49 & 47.87 \\
% \bottomrule
% \end{tabular}
% \vspace{-0.2cm}
% % \caption{Comparison of different denoising methods. We mark the \textbf{best} and \underline{second best} results.}
% \caption{Ablation about time window number. }
% \label{tab:ablation2}
% \vspace{-0.6cm}
% \end{table}

\noindent \textbf{Effects of Time Windows Number.} 
% We evaluate model performance with time windows of 3, 5, and 8 on synthetic data. Increasing the number of time windows improves the model's ability to capture temporal dependencies, with a \todo{1.5\%} improvement from 3 to 5 windows and a \todo{0.2\%} improvement from 5 to 8. However, the inference time increases from \todo{X to Y} when going from 5 to 8 windows. Beyond this point, further increases in time windows show diminishing returns, indicating an optimal balance between temporal depth and model complexity.
% We evaluate model performance with time windows of 3, 5, and 8 on synthetic data. Increasing the number of time windows improves the model's ability to capture temporal dependencies, with a \todo{1.5\%} improvement from 3 to 5 and \todo{0.2\%} from 5 to 8. However, the inference time increases from \todo{X to Y} when going from 5 to 8 windows. Beyond this, further increases in time windows yield diminishing returns, emphasizing the balance between temporal depth and model complexity.
In Table~\ref{tab:ablation2}, we evaluate model performance with time windows of 3, 5, and 8 on validation set. Increasing the number of time windows improves the model's ability to capture temporal dependencies, with a 9.0\% improvement in DA from 3 to 5 windows and an additional 3.8\% improvement from 5 to 8.  However, the inference speed decreases by 30\% when the window size is increased from 3 to 5, and by 48\% when increased from 3 to 8. Therefore, we set the number of time windows to 5, as further increases in time windows yield diminishing returns, emphasizing the need to balance temporal depth and computational efficiency.
% \noindent \textbf{Effects of }
\vspace{-0.1cm}
\subsection{Generalization to Snowy Datasets}
\vspace{-0.1cm}
% \begin{figure}[t]
%   \centering
%    \includegraphics[width=1.0\linewidth]{ICCV2025-Author-Kit-Feb/img/snow_v6.png}
%   \caption{Qualitative results on snowy datasets.}
% \label{fig:snow}
% \end{figure}

While primarily designed for event-based deraining, our method also generalizes well to snow removal, as demonstrated in Figure~\ref{fig:realworld} (last two rows). Without architectural changes, it delivers promising results under snowy conditions, showcasing robustness to diverse weather patterns. Additional visual results are included in the Appendix.
\vspace{-0.2cm}
\section{Conclusion}
\vspace{-0.1cm}
\label{sec:conclusion}
To the best of our knowledge, \model\ is the first point-based framework for event camera deraining. It adopts a 4D event cloud to preserve native temporal resolution via inter- and intra-temporal windows, a spatiotemporal decoupling and fusion module for efficient feature extraction, and a multi-scale state space model that captures rain dynamics across spatial and temporal scales with linear complexity. A frequency regularization term and cross-entropy loss guide learning, and we contribute a large-scale dataset with diverse rain scenarios. Extensive experiments demonstrate SOTA performance on synthetic and real-world data. The method is efficient, lightweight, and real-time capable, with potential for robust multi-agent perception~\cite{chen2024ddl, chen2024soscheduler}.
\\
\noindent\textbf{Acknowledgments} 
% This paper was supported by the National Key R\&D program of China No. 2022YFC330
% 0703, Natural Science Foundation of China under Grant 62371269, Guangdong Innovative and Entrepreneurial Research Team Program (2021ZT09L197), Meituan Academy of Robotics Shenzhen.
This paper was supported by the Natural Science Foundation of China under Grant 62371269, Guangdong Innovative and Entrepreneurial Research Team Program (2021ZT09L197), and Meituan Academy of Robotics Shenzhen.

{
    \small
    \bibliographystyle{ieeenat_fullname}
    \bibliography{main}
}

\end{document}